**Development of Statewide AADT Estimation Model from Short-Term Counts: A Comparative Study for South Carolina**

**Sakib Mahmud Khan***
**Ph.D. Candidate**
Clemson University
Glenn Department of Civil Engineering
351 Fluor Daniel Engineering Innovation Building, Clemson, SC 29634
Tel: (864) 569-1082, Fax: (864) 656-2670
Email: sakibk@g.clemson.edu

**Sababa Islam**
**Ph.D. Student**
139 Lowry Hall, Glenn Department of Civil Engineering, Clemson University
Clemson, South Carolina 29634
Tel: (864) 633-9056, Fax: (864) 656-2670
Email: sislam@clemson.edu

**MD Zadid Khan**
**Ph.D. Student**
Glenn Department of Civil Engineering, Clemson University
351 Fluor Daniel Engineering Innovation Building, Clemson, SC 29634
Tel: (864) 359-7276, Fax: (864) 656-2670
Email: mdzadik@clemson.edu

**Kakan Dey, Ph.D.**
**Assistant Professor**
Civil and Environmental Engineering
West Virginia University
ESB 647, 401 Evansdale Drive, PO Box 6070, Morgantown, West Virginia
Tel: (304) 293-9952   Fax: (304) 293-7109
Email: kakan.dey@mail.wvu.edu

**Mashrur Chowdhury, Ph.D., P.E., F.ASCE**
**Eugene Douglas Mays Endowed Professor of Transportation and**
**Professor of Automotive Engineering**
Clemson University
Glenn Department of Civil Engineering
216 Lowry Hall, Clemson, South Carolina 29634
Tel: (864) 656-3313   Fax: (864) 656-2670
Email: mac@clemson.edu

**Nathan Huynh, Ph.D.**
**Associate Professor**
Department of Civil and Environmental Engineering
University of South Carolina
300 Main Street, Columbia, SC 29208
Tel: (803) 777-8947   Fax: (803) 777-0670
Email: huynhn@cec.sc.edu

*Corresponding author

Abstract: 258 words + Text: 4660 + References: 552 + 6 tables/figures: 1500 = **6970** words

**ABSTRACT**

Annual Average Daily Traffic (AADT) is an important parameter used in traffic engineering analysis. Departments of Transportation (DOTs) continually collect traffic count using both permanent count stations (i.e., Automatic Traffic Recorders or ATRs) and temporary short-term count stations. In South Carolina, 87% of the ATRs are located on interstates and arterial highways. For most secondary highways (i.e., collectors and local roads), AADT is estimated based on short-term counts. This paper develops AADT estimation models for different roadway functional classes with two machine learning techniques: Artificial Neural Network (ANN) and Support Vector Regression (SVR). The models aim to predict AADT from short-term counts. The results are first compared against each other to identify the best model. Then, the results of the best model are compared against a regression method and factor-based method. The comparison reveals the superiority of SVR for AADT estimation for different roadway functional classes over all other methods. Among all developed models for different functional roadway classes, the SVR-based model shows a minimum root mean square error (RMSE) of 0.22 and a mean absolute percentage error (MAPE) of 11.3% for the interstate/expressway functional class. This model also shows a higher R-squared value compared to the traditional factor-based model and regression model. SVR models are validated for each roadway functional class using the 2016 ATR data and selected short-term count data collected by the South Carolina Department of Transportation (SCDOT). The validation results show that the SVR-based AADT estimation models can be used by the SCDOT as a reliable option to predict AADT from the short-term counts.

## INTRODUCTION

Annual average daily traffic (AADT) is used in many transportation engineering projects (i.e., roadway design, transportation planning, traffic safety analysis, highway investment decision making, highway maintenance, air quality compliance studies, and travel demand modeling). The accuracy of AADT estimation for short-term count stations is critical for any transportation projects that use AADT as an input parameter. For example, AADT is a vital input variable for the Safety Analyst software and the Highway Safety Manual *(1)*. Moreover, as a part of the traffic monitoring program, every state Department of Transportation (DOT) has to report the statewide estimated AADT to Federal Highway Administration annually *(2)*.

Using permanent traffic count stations or Automatic Traffic Recorder (ATR), AADT can be directly measured for any ATR location. An ATR collects traffic data 24 hours per day for the whole year using traditional inductive loops, microwave radar sensors, magnetic counters, and piezoelectric sensors. However, the installation of ATRs at thousands of traffic count stations to count vehicles over the year is not economically feasible. For this reason, ATRs are installed at a limited number of strategic locations. To supplement these ATRs, short-term traffic counts (i.e., 24/48-hour counts) are performed and the data collected at these locations are used for AADT estimations. The data collection frequencies at short-term count stations are inconsistent among different states. While short-term counts are collected annually in some states, others collect every few years *(3)*. Traditionally, for the locations that do not have any ATRs, the AADT is estimated using expansion factors (i.e., seasonal, daily, monthly, growth and axle factors). Many DOTs, such as the SCDOT, use this method to determine AADT from short-term counts. The expansion factors are calculated based on the continuous traffic volume data collected from the permanent count stations *(4)*. To develop accurate expansion factors, permanent and short-term count stations are combined based on the roadway functional class and geographical locations *(5)*. After grouping, permanent count station data are used to determine the expansion factors. Data from short-term count stations within the same group are used to determine AADT based on the short-term count locations by applying these factors. This method of AADT estimation at short-term count stations is known as the factor-based method, which has some drawbacks. There are no defined guidelines or established standards about assigning the expansion factors from ATR to short-term traffic count stations *(3)*. Moreover, the relatively small number of ATRs in the lower roadway functional classes makes it challenging for the development of accurate expansion factors for the larger number of short-term count stations on local roads. One solution is to have more permanent count stations in the lower functional classes, but that has a significant investment requirement. To address this issue, several models have been developed in this study based on machine learning and regression. More specifically, the study objective is to develop AADT estimation models which can be used by DOTs to reliably estimate AADT for the short-term count stations. The developed model was compared against factor-based method used by DOTs to estimate AADT. The following sections discuss the related studies, research contribution, method and findings from the developed AADT estimation models.

## LITERATURE REVIEW

In this section, previous work related to AADT estimation methods have been reviewed. Among all methods, regression analysis is the most popular method for AADT estimation. Xia et al. *(6)* considered roadways characteristics in AADT estimation in Florida. Zhao and Chung *(7)* used GIS to extract land-use and accessibility information to be used in regression models. Zhao and Park *(8)* estimated regression parameters locally (i.e., based on observations near the estimation location) instead of globally and referred to it as Geographically Weighted Regression (GWR) method. The comparison showed that GWR is more accurate than ordinary linear regression. Jiang et al. *(9)* proposed to use the weighted average of (i) the count from the growth factor-based method, which uses last years' data to predict AADT, and (ii) the traffic count from the current year's image containing traffic information. Kingan and Westhuis *(10)* proposed a more robust regression method, which minimizes a proportion of the sum of the smallest squared residuals for AADT estimation since the ordinary least square method is vulnerable to outliers. Yang et al. *(11)* studied variable selection and parameter estimation using different groups of variables. Significant variables can be identified by the smoothly clipped absolute deviation penalty method, which can also determine regression coefficients.

Among the machine learning techniques, Artificial Neural Network (ANN) has been used extensively in studying driver behavior, pavement maintenance, vehicle classification, traffic forecasting and pattern analysis *(12)*. Moreover, a well-trained ANN model can model the complex relationship of hourly traffic volume data, finding the traffic pattern and estimating the AADT without grouping the permanent and short-term count stations *(13)*. Neural networks have also been used to determine AADT using short-term traffic counts *(3, 13)*. Support vector regression (SVR) has also been used for similar applications, specifically for predicting and comparing travel times with the base-line travel time prediction using real-time traffic data. A study by Lin indicated that SVR has greater learning potential than ANN *(14)*. However, limited research has been conducted using SVR in traffic data analysis *(15)*. In the exploration of the potential of SVR for short-term traffic speed predictions, Vanajakshi et al. found that when training data is limited SVR performed better then ANN, although further study is required to determine the advantages and disadvantages of both methods *(15)*. Furthermore, in their evaluation of the performance of a modified SVR, which is SVR with data-dependent parameters (SVR-DP), Castro-Neto et al. collected AADT values from 1985 and 2004 from Tennessee *(16)*. A subsequent comparison of the SVR-DP approach with the popular Holt-Winters exponential smoothing (HW method) and the ordinary least-squares (OLS) linear regression methods showed that the SVR-DP outperformed both.

## METHOD

This section outlines the five steps followed for AADT estimation model development and evaluation applying all three techniques- ANN, SVR and OLS regression. The first three steps include data collection, data processing, and feature selection. Step 4 is the model development step and step 5 is the model evaluation step.

## Step 1: Data Collection
### ATR Data
The South Carolina DOT maintains a total of 164 permanent count stations (i.e., ATR) on different functional classes, which include 83 stations on interstates, 59 on arterials, 15 on collectors and seven on local roads. An interactive web-scraping model, developed in Python using a library called Selenium, is used to collect data from the SCDOT ATR data reporting website. ATRs with more than six months of missing data are not used. Data for a day are not used if any hourly volume data for that day is missing, which can be caused by ATR data collection equipment hardware and/or software malfunctions *(17)*. FIGURE 1 shows the number of ATRs with missing count information. Among the 112 ATRs, the figures shows that 98 ATRs had less than one month of missing data, while 14 ATRs had missing data for more than one month.

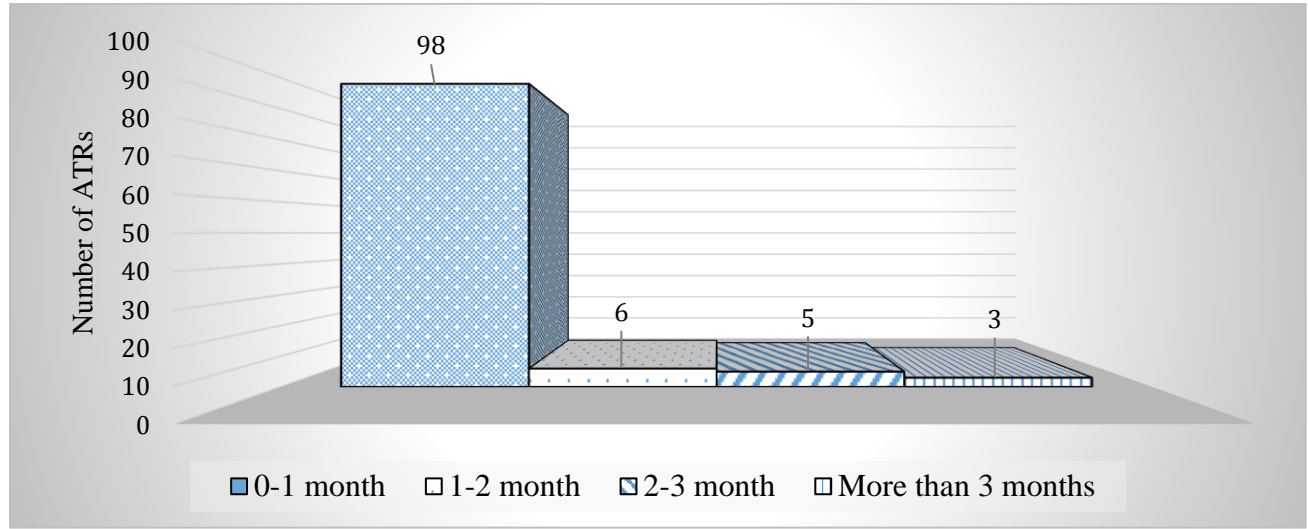

**FIGURE 1**      Number of ATRs with Missing Counts

Data are collected for multiple years (i.e., for year 2011 and year 2016) to validate the robustness of the models.

*Socio-economic Data*

In addition to the ATR data, the authors have collected the following socio-economic information from the US Census Bureau data: (i) income, (ii) employment, (iii) percent below poverty, (iv) number of vehicles, (v) urban or rural, and (vi) number of housing units *(18)*.

## Step 2: Data Preprocessing

After data are collected, they are prepared for input into the models. The variables are divided into four groups, (i) traffic volume features, (ii) socio-economic features, (iii) roadway characteristics and (iv) categorical features. For the ANN and SVR models, variables of categories (i), (ii) and (iv) are considered. For the regression models, variables of categories (i), (ii) and (iii) are considered. For the ANN and SVR models, data are divided into training and test sets. Two-thirds of the data are used as training, the rest are used to test the models.

*Traffic Volume Features*

To develop the AADT estimation models using the machine learning techniques, the authors have followed the study conducted by Sharma et al., in which ANN is used to develop an AADT estimation model for Minnesota *(3)*. However, rather than using 48 hourly volume factors in the Minnesota model, the authors have used 24 hourly volume factors in this study. The equation for developing the hourly volume factor is expressed below:

$$Hourly\ volume\ factor = \frac{Traffic\ volume\ for\ an\ hour\ (e.g.\ traffic\ volume\ for\ 7AM-8AM\ on\ monday\ of\ january, 2011)}{Sum\ of\ 24\ hourly\ volume\ of\ that\ day} \quad (1)$$

For regression based models, hourly volume data are directly used to estimate AADT, rather than converting them to factors.

*Socio-economic Feature*

The zip-code level data are processed using ArcGIS. The geographic map of the traffic counts, roadway characteristics, and the zip-codes are joined and the desired data are extracted for each location.

*Roadway Characteristics*

Roadway characteristics are the most widely used variables in AADT estimation. In this research, only one roadway characteristic is considered, the functional classification. Roadways from a higher functional class with a higher number of lanes always have the higher traffic volume. According to ATR stations' roadway functional classification, we considered three groups of functional classes for the model development step: i) interstates and expressways, ii) principal and minor arterials and (iii) all ATR models.

*Categorical Features*

Most AADT estimation models only use hourly volume data (continuous features/variables) *(3, 13)*, no categorical features/variables are used along with the continuous hourly volume data. In this study, however, the authors have developed models with continuous (hourly volume) and categorical features. The categorical features that have been considered are: (i) day of week and (ii) month of year. The reason for considering these two variables is that they have a significant impact on traffic volumes. Monday traffic volume in January is very different from Saturday traffic volume in October. Binary variables are used for quantifying these categorical features. For example, to develop the day of week variables, one feature is developed for each day of the week for a total of seven features. If a data point is for Monday, then the Monday feature is assigned the value 1, and the features for the other days of the week are assigned 0. A similar method is used to develop the month of the year.

*AADT Factors*

The target feature used in this study is a factor of the actual AADT called the AADT factor. For each day, the AADT factors are calculated using Equation 2.

$$AADT\ factor = \frac{AADT}{Sum\ of\ 24\ hourly\ volume\ of\ that\ day} \quad (2)$$

AADT is estimated by using 24 hourly volume data for each day. For each ATR, the AADT is computed by calculating a simple average mean of the estimated AADT from each day of the year as mentioned in the traffic monitoring guide *(9)*.

## Step 3: Feature Selection

Feature selection for ANN and SVR models is performed to reduce the use of irrelevant data in developing predictive models and to improve the model performance in terms of speed and accuracy *(10)*. The sequential feature selection method is used to identify the desired features *(19)*. It is a simple greedy search method starting with an empty set of features and sequentially adding the most impactful feature in each step until the desired result from the criterion function is achieved (*11, 17*). In this study, the combination of features with the least residual sum of square error for predicting target values and features are chosen. The sequential feature selection method is applied for the 24 hourly volume data, and 20 hourly volume data are considered finally. For the regression models, stepwise regression method was used to identify the most significant features to estimate AADT. The correlation between the independent variables were checked and no independent variables were found to be correlated.

## Step 4: Model Development

Once the hourly volume features are selected, a combination of the hourly volume features with the categorical features and socio-economic features is used for developing the ANN and SVR models, where the output is the AADT factor. For the regression models, the dependent variable is the AADT. The hourly volume, socio-economic factors and roadway characteristics are the independent variables. At first, three major models were developed for ANN and SVR: (i) the interstate and expressway model, (ii) the principal and minor arterial model and (iii) the general model or all ATR model which can be applied to any roadway functional class. The input data set is filtered accordingly, since the models are functional class specific. Depending on the combinations of hourly volume factors, socio-economic and categorical features, nine alternatives are created for each functional class model. This is done to determine which combination of the features produce the best model. FIGURE 2 shows the combination of features for each alternative.

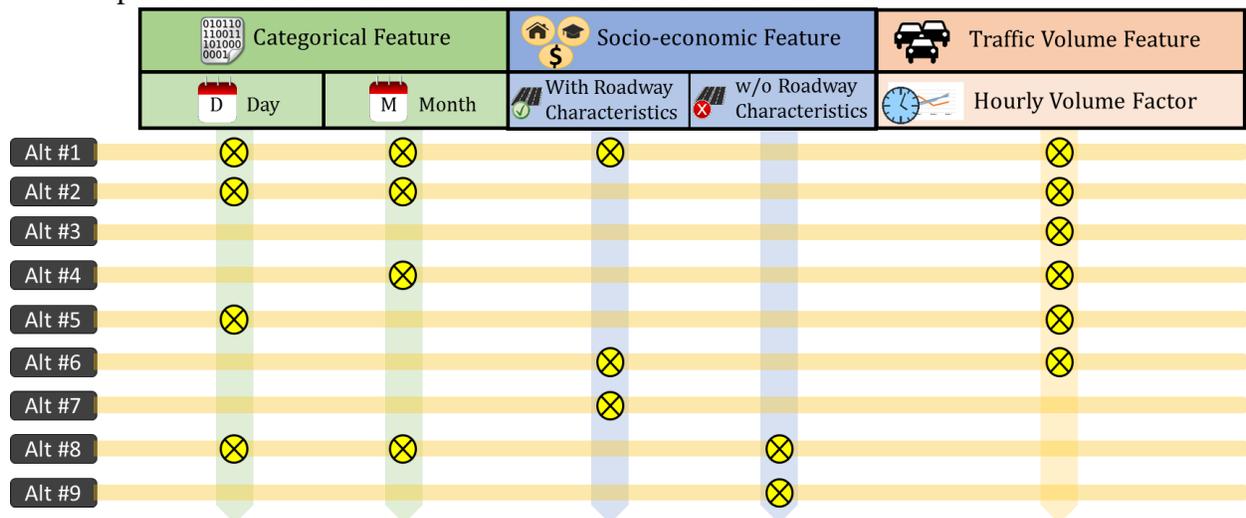

**FIGURE 2**    Model Alternatives to Estimate AADT

The interstate and expressway model and principal and minor arterial model are developed using the permanent traffic count stations available for the specific roadway functional class. On the other hand, for the regression model, the functional class information is used as an input, so only one general purpose model is developed.

The all-ATR model uses data from all available permanent traffic count stations. The motivation for developing this model is to create a general model that is not roadway functional class specific, and is applicable for estimating AADT from a short-term count station on any functional class. Depending on the combination of the training features, the all ATR model has only four alternatives instead of nine, alternative 2, 3, 4 and 5. This is because the alternatives consisting of socio-economic features are not used, since that data are not available for all ATRs. The following sections discuss model development using ANN, SVR and OLS regression.

*Artificial Neural Network (ANN)*

In this study, the authors used a multilayered, feed-forward, backpropagation neural network model for supervised learning. The developed neural network model has three layers: the input, hidden and output layer. This ANN model is a feed-forward network as it feeds the output of one layer to the next layer. A tan-sigmoid transfer function is used for calculating the output from each neuron. One of the remarkable characteristics of a back-propagation neural network is its ability to propagate the effects of error backward through the network after every training case *(20)*; hence this algorithm is chosen for estimating AADT in this study. The training algorithm selected is the Levenberg-Marquardt, which is recommended for most of the prediction problems unless the data set is noisy and small *(21)*.

In this study, the authors built different ANN models with a different number of hidden neurons and chose the number of neurons which provides the least root mean square error (RMSE) for the related models. The number of hidden neurons used in this study varied based on the different models used. The neural network model was developed using the MATLAB library function NNtool.

*Support Vector Regression (SVR)*

SVR is used for nonlinear regression by mapping the training set onto a higher dimensional, kernel-induced feature space. In this study, a SVR algorithm with radial basis kernel function is chosen and implemented using the MATLAB LIBSVM library tool *(22)*. The cost function (C) is used to determine the tradeoff between the complexity of the model and degree of deviation from $\varepsilon$ (epsilon) that canbe tolerated. The $\varepsilon$ parameter controls the width of the $\varepsilon$ insensitive zone, and is used to fit the training data *(22)*. The model relies on a single subset of the training samples; the cost function ignores the training samples within the $\varepsilon$ tube (i.e., a certain threshold distance from the prediction). The initial value of C and $\gamma$ (gamma) are determined based on the grid search method with a 5-fold cross validation. Cross validation is performed to reduce the bias of a training dataset on the model parameters. The trial and error method is used to find the parameters that yield the least error on AADT prediction. C and $\gamma$ values are used as powers of two, where the range of powers for C values are $-3$ to 15, and the range of powers for $\gamma$ values are $-15$ to 3, as suggested in *(22)*. The value of the parameters varied from model to model with the change in training data.

*Regression*

The ordinary least squares (OLS) regression method was used to develop the model. OLS is a method used to determine the unknown parameter in a linear regression model. The dependent variable in the regression model is AADT. Hourly volume data, roadway characteristics and socioeconomic characteristics are used as independent variables for the models.

## Step 5: Model Evaluation

Once the alternatives for each AI model have been developed, the results are compared against the ground truth AADT from the ATR data to calculate the model accuracy. Based on the accuracy of the alternatives, a final alternative for each functional class and a general model are selected as the best models. For regression, three models are developed, one of which is the general model.

The performance of the alternatives is assessed based on the root mean square error (RMSE) and mean average percentage error (MAPE). The formula used for RMSE and MAPE are shown below as Equations (3) and (4).

$$\text{RMSE} = \sqrt{\frac{\sum_{i=1}^{n}(Y-yi)^2}{n}} \qquad (3)$$

$$\text{MAPE} = \frac{1}{n}\sum\frac{|Y-yi|}{Y}*100 \qquad (4)$$

where $Y$ = actual value, $y_i$ = predicted value for $i$-th observation, and $n$ = number of observations The best alternative for each model is selected based on RMSE and MAPE. Then, the best alternatives for each functional class are compared against the regression-based models.

After that, the robustness and effectiveness of the selected models are validated in two steps. First, the models are validated using a different input dataset, such as ATR data from a different year. The whole model development process is followed again, but with the ATR data from a different year. This step is very important to prove the transferability of the model. If the performance of the models are satisfactory, then the AADT estimation models can be used for any future year.

Second, the authors have used short-term counts as input to the model and predict the number of AADT short counts for that roadway. The predicted AADT is compared with the AADT calculated from the nearby ATR data, which represents the ground truth data in this case. If the prediction error is reasonable, then it can be concluded that the models perform well in predicting AADT from short-term counts, and they can be used by DOTs for statewide estimation of AADT from short-term counts.

## ANALYSIS AND RESULTS

As a case study, 2011 is chosen as the year for all data collection for model development; 2016 data are used for machine learning-based model validation. From the 2011 ATR data, only 112 ATRs are used in the model development. Other ATRs had insufficient data (e.g., missing more than six months of data) for accurate AADT estimation. Socio-economic data are obtained from the 2011 census data.

## AADT Estimation with Machine Learning Techniques

Using the input data from 2011, model alternatives are developed. Then, the results from ANN and SVR models are summarized to determine which alternatives perform better based on two performance measures, RMSE and MAPE. The alternatives with the least error is identified and then compared against the regression model and factor-based model. After this comparison, the best alternatives for ANN and SVR with the lowest RMSE and MAPE are carried forward to the evaluation step, where the alternatives are validated using a new dataset from the year 2016. TABLE 1 summarizes the results of the ANN and SVR models.

### Model 1: Interstate & Expressway Models

The RMSE and MAPE values for different alternatives of the Interstate & Expressway functional class are shown in TABLE 1. Each alternative consists of different combinations of training features such as traffic volume factors, socio-economic variables, and other categorical features (day of week, month, urban/rural roadway). The performances of the SVR alternatives are better than those of ANN alternatives. The best alternative found from this comparison is the SVR alternative 2 (Day, Month, and hourly volume factors as features), with an RMSE of 0.22 and MAPE of 11.3%. A comparison of the alternative errors shows that the addition of socio-economic features lowers the model performance.

**TABLE 1** ANN and SVR Alternative Models' Performance

| Model | Model Alternative | Performance in terms of RMSE and MAPE | | | |
|---|---|---|---|---|---|
| | | ANN | | SVR | |
| | | RMSE | MAPE (%) | RMSE | MAPE (%) |
| Interstate and Expressway Model | Alternative 1 | 0.37 | 17.6 | 0.37 | 16.7 |
| | Alternative 2 | 0.25 | 11.9 | 0.22 | 11.3 |
| | Alternative 3 | 0.26 | 13.8 | 0.23 | 12.7 |
| | Alternative 4 (Monday) | 0.27 | 12.5 | 0.23 | 11.5 |
| | Alternative 5 (January) | 0.52 | 17.8 | 0.26 | 11.7 |
| | Alternative 6 | 0.48 | 27.0 | 0.37 | 16.6 |
| | Alternative 7 | 0.44 | 25.4 | 0.37 | 16.7 |
| | Alternative 8 | 0.38 | 19.8 | 0.38 | 19.8 |
| | Alternative 9 | 0.37 | 19.5 | 0.37 | 16.7 |
| Principal and Minor Arterial Model | Alternative 1 | 1.19 | 51.5 | 0.34 | 17.3 |
| | Alternative 2 | 0.29 | 17.4 | 0.29 | 13.7 |
| | Alternative 3 | 0.32 | 18.7 | 0.31 | 14.1 |
| | Alternative 4 (Monday) | 0.46 | 12.5 | 0.32 | 11.5 |
| | Alternative 5 (January) | 0.53 | 17.8 | 0.35 | 11.7 |
| | Alternative 6 | 2.09 | 85.6 | 0.35 | 16.5 |
| | Alternative 7 | 2.15 | 85.6 | 0.35 | 17.6 |
| | Alternative 8 | 0.41 | 21.7 | 0.34 | 16.5 |
| | Alternative 9 | 2.60 | 124.4 | 0.35 | 17.2 |
| All-ATR model | Alternative 2 | 0.36 | 17.4 | 0.35 | 15.1 |
| | Alternative 3 | 0.32 | 15.9 | 0.31 | 13.7 |
| | Alternative 4 | 0.32 | 15.6 | 0.46 | 18.4 |
| | Alternative 5 | 0.31 | 14.8 | 0.33 | 13.6 |

*Model 2: Principal/Minor Arterial Model*

The RMSE and MAPE of the different SVR and ANN alternatives for principal/minor arterials are shown in TABLE 1. The results are very similar to Model 1 results. For the principal/minor arterial functional class of roadways, SVR alternatives also performs better than ANN alternatives. Moreover, like the interstate and expressway model, the addition of the socio-economic variables lowers the performance of the models. The SVR alternative 2 has the lowest RMSE (0.29) and MAPE (13.7%).

*Model 3: All-ATR Model*

The RMSE and MAPE of different SVR and ANN alternatives for all 112 permanent count stations on highways in South Carolina are summarized in TABLE 1. The training features used for developing these alternatives are the hourly volume factors, month of year and day of week. The socioeconomic data are not available for all permanent count locations; therefore, not all alternatives are developed for this model. A comparison of the results for the different alternatives shows that SVR alternative 3 produces the lowest RMSE (0.31) and MAPE (13.7%). Alternative 3 uses only the hourly volume data for all available permanent count stations as input (no other variables are used). This general all-ATR model provides a reasonable estimate of AADT considering the variability of the training dataset.

## AADT Estimation with Regression Models

Three regression models are developed.  One for expressways/interstates, one for principal/minor arterials, and one for all functional classes. The variables and R-squared values of these model are shown in TABLE 2.

**TABLE 2**,   Summary of Regression Models

| Model | Variable | R-squared Value |
|---|---|---|
| Interstate and expressway model | • Traffic volume features: Count for 1-4,  6-7, 9-14, 16, 18-24 hour count<br>• Roadway characteristics: Functional Class<br>• Socio-economic features: Urban Area, Income, Employment, Person below poverty, Vehicles, Housing Unit | 0.884 |
| Principal and minor arterial | • Traffic volume features: Count for 1-5, 7-8, and 10-22  hour count<br>• Roadway characteristics: Functional Class<br>• Socio-economic features: Urban Area, Income, Employment, Person below poverty, Vehicles, Housing Unit | 0.973 |
| All ATR | • Traffic volume features: $1-5$, $7 - 8$, $10 - 16$, $18-20$ and 22 hour count<br>• Roadway characteristics: Functional Class<br>• Socio-economic features: Urban Area, Income, Person Below Poverty, Vehicles | 0.924 |

## Comparison between SVR and Regression-based AADT Estimation Model

To compare the effectiveness of the SVR models, the best alternative for SVR is compared against the results of the traditional regression models. For this comparison, the same ATRs, which were used for testing ANN and SVR-based AADT estimation models, are used. In total, estimated AADT from 30 ATRs (with 351 days for each ATR) are compared. The MAPE (%) is calculated for all three models (i.e., interstate, arterial and all-ATR model) to compare the performance of SVR and regression. The results of the comparison are presented in FIGURE 3. Based on this comparison, it can be concluded that the SVR alternative provides more accurate AADT estimation for different functional classes.

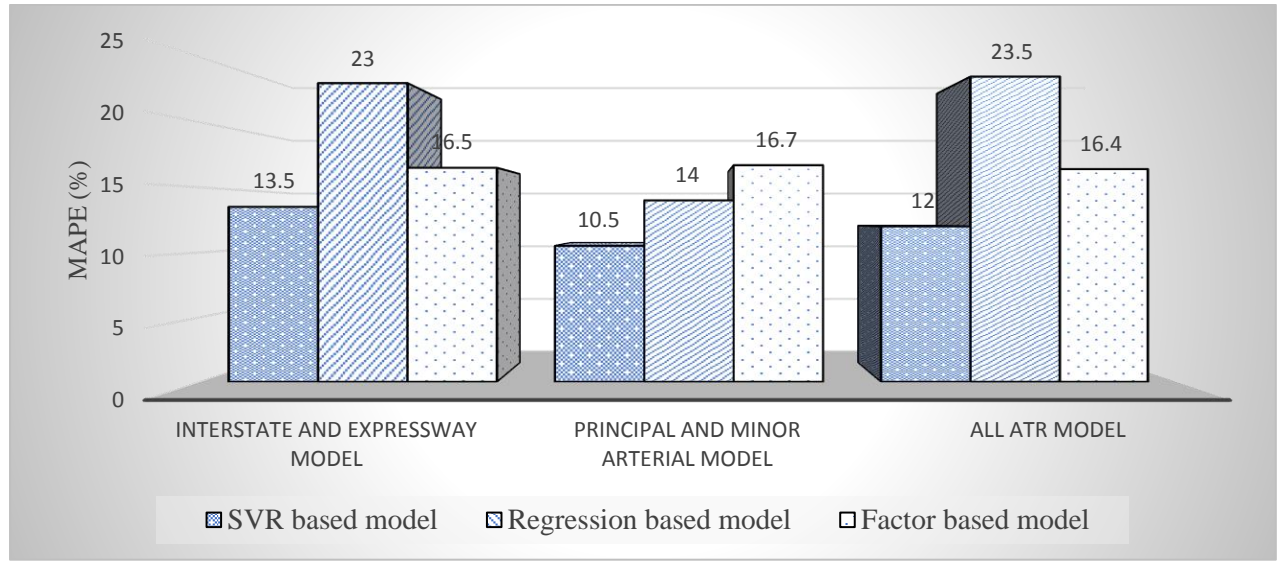

**FIGURE 3** MAPE Values for AADT Estimation Models

## Comparison between SVR and SCDOT AADT Estimation (Factor) Method

One of the objectives of this study is to find the efficacy of the models developed using the machine learning techniques over the traditional factor-based method currently used by SCDOT. In the traditional factor-based method for estimating AADT, SCDOT uses two types of factors: 1) seasonal or monthly factors, and 2) axle correlation factors. These factors are gathered for each of the roadway functional classes. The short-term counts conducted in these functional classes are multiplied with these functional class specific factors to estimate AADT. To compare the AADTs estimated by SVR and factor-based method, the same ATRs, which were used for testing ANN and SVR-based AADT estimation models are used. In total, estimated AADTs from 30 ATRs (with 351 days for each ATR) are compared. In the SVR-based model, the estimated AADT factors are multiplied by the sum of 24 hourly volumes to calculate the AADT. To estimate AADT using the factor-based method, the sum of 24 hourly volume for the selected day is multiplied by the monthly factor and axle correction factor. In this section, the AADT values are compared for the interstate and expressway model, principal/minor arterial model and all ATR models. The MAPE values for the three models are presented in FIGURE 3. It can be observed that SVR yields a better MAPE value for the principal and minor arterial models than the interstate and expressway model. For the principal and minor arterial model, the MAPE value of the SVR model is 10.5% which is lower than the MAPE value of the traditional factor-based method (16.7%).

## SVR Model Validation

The SVR model (as depicted in Figure 2) is superior for all functional classes. In this section, SVR-based models are validated using two methods. First, the AADT estimation model is developed with the 2016 ATR data, instead of 2011 ATR data. The only input features required for alternative 2 are 24 hourly volumes and categorical features. The 2016 ATR data is collected from all 164 ATRs. Model alternative 2 is trained with $2/3^{rd}$ of the 2016 ATR data, and tested with $1/3^{rd}$ of the 2016 ATR data. Alternative 3 is also developed for all ATR models, since this was the best alternative for all ATR models using the 2011 dataset. The results are shown in the TABLE 3. Alternative 2 performs better than alternative 3 for the all-ATR model. This represents a change from the 2011 data. As an example, the RMSE values for 2011 and 2016 for alternative 2 of the principal and minor arterial models are 0.29 and 0.31, respectively. The MAPE values for 2011 and 2016 for alternative 2 of the principal and minor arterial models are 13.7% and 11.7%, respectively. These results indicate that the models are robust and stable despite them being developed using datasets from two different years. Based on this validation, it can be concluded that alternative 2 is the best alternative for all three models (i.e., interstate and expressway model, principal and minor arterial model and the all-ATR model).

The second method used to validate the developed models is to use short-term count data collected from five locations near existing ATRs on interstates in 2016. The ATR IDs of those locations are 98, 145, 102, 50, and 110. The authors have used this data as input to alternative 2 for both interstate and expressway model and all ATR model, since these are the models that can be used for interstates. The training dataset for SVR includes all ATRs except for the five ATRs closest to the short-term count locations. The results of the model validation are shown in TABLE 3. The interstate and expressway model performs slightly better than all-ATR model in terms of RMSE and MAPE.

**TABLE 3**   SVR Model Validation Using ATR data and Short-Term Count Data

| Validation Data | Model | RMSE | MAPE (%) |
|---|---|---|---|
| 2016 ATR data | All ATR model (Alternative 2) | 0.53 | 11.9 |
| | All ATR model (Alternative 3) | 0.55 | 12.7 |
| | Interstate and expressway model (Alternative 2) | 0.61 | 12.8 |
| | Principal and minor arterial model (Alternative 2) | 0.31 | 11.7 |
| Short-term count (48 hour volume data, 5 locations) | All ATR model (Alternative 2) | 0.11 | 10.2% |
| | Interstate and expressway model (Alternative 2) | 0.09 | 9.5% |

## CONTRIBUTIONS OF THIS RESEARCH

This study introduces several new dimensions to the literature on AADT estimation. The comparisons made between the machine learning-based model, regression model and factor-based model (currently used by SCDOT) is the first such application. Furthermore, this study investigated multiple roadway functional classes separately, and subsequently developed models and identified the best model for each functional class. Lastly, this is the first study to develop a practical implementation of a more reliable, statewide AADT estimation model for the state of South Carolina.

**CONCLUSIONS**

In this study, the AADT estimation models are developed to estimate AADT at the short-term count stations on different roadway functional classes in South Carolina. The AADT estimation models are created using two machine learning methods (i.e., ANN, SVR) and ordinary least squares regression methods. This study reveals that the AADT estimation model that use SVR outperformed the other models. The best SVR model uses hourly volume data as well as day of week and month of year categorical features in estimating AADT for short-term count stations. Adding other features such as socio-economic factors lowers the performance of the model. This finding suggests that the AADT of a location depends primarily on the traffic patterns (i.e., day of week, month of year). The developed model can be incorporated into an AADT estimation toolkit to be used by DOTs to predict AADT for any location using short-term counts.

**ACKNOWLEDGEMENT**

The authors acknowledge the South Carolina Department of Transportation, which provided funding for this research.

**DISCLAIMER**

The contents of this report reflect the views of the authors who are responsible for the facts and the accuracy of the presented data. The contents do not reflect the official views of SCDOT or FHWA. This report does not constitute a standard, specification, or regulation.